\documentclass{article}

\PassOptionsToPackage{numbers,compress}{natbib}
\usepackage[position, final]{neurips_2026}

\usepackage[utf8]{inputenc} % allow utf-8 input
\usepackage[T1]{fontenc}    % use 8-bit T1 fonts
\usepackage{hyperref}       % hyperlinks
\usepackage{url}            % simple URL typesetting
\usepackage{booktabs}       % professional-quality tables
\usepackage{amsfonts}       % blackboard math symbols
\usepackage{amsmath}        % \text, align, etc.
\usepackage{amssymb}        % \checkmark and other symbols
\usepackage{nicefrac}       % compact symbols for 1/2, etc.
\usepackage{microtype}      % microtypography
\usepackage{xcolor}         % colors
\usepackage[most]{tcolorbox}
\usepackage{enumitem}
\usepackage[framemethod=default]{mdframed}

\usepackage[capitalize,noabbrev]{cleveref} 
\usepackage{booktabs, tabularx}

\title{Sampling Is Not Curiosity: \\
Why LLM Agents Should Investigate
}

\author{%
  Alfonso Amayuelas$^{1,2}$ \quad
  Piotr Piękos$^{3}$ \quad
  Xin Eric Wang$^{1}$ \\
  \bfseries
  William Yang Wang$^{1}$ \quad
  Jürgen Schmidhuber$^{3}$ \quad
  Alex Pentland$^{2}$ \\[4pt]
  \normalfont
  $^{1}$University of California, Santa Barbara \quad
  $^{2}$Stanford University\\
  $^{3}$King Abdullah University of Science and Technology\\[4pt]
  \normalfont\small
  Correspondence: \texttt{amayuelas@ucsb.edu}
}

\newtcolorbox{mybox}[1]{
  enhanced,
  colback=white,
  colframe=black,
  boxrule=1pt,
  arc=4pt,
  fonttitle=\bfseries,
  coltitle=white,
  colbacktitle=black,
  title=#1,
  left=10pt, right=10pt, top=8pt, bottom=8pt,
  toptitle=2pt, bottomtitle=2pt,
  lefttitle=10pt,
  titlerule=0pt,
}

\newtcolorbox{callout}{
  enhanced,
  breakable,
  colback=gray!8,           % subtle background tint
  colframe=black,            % left rule color
  boxrule=0pt,               % no full border
  frame hidden,
  borderline west={2pt}{0pt}{black},  % left vertical bar
  sharp corners,
  left=10pt, right=10pt, top=8pt, bottom=8pt,
}

\newmdenv[
  topline=false,
  rightline=false,
  bottomline=false,
  linewidth=2pt,
  linecolor=black!70,
  innerleftmargin=12pt,
  innerrightmargin=0pt,
  innertopmargin=4pt,
  innerbottommargin=4pt,
  skipabove=8pt,
  skipbelow=8pt,
  leftmargin=0pt,
  rightmargin=8pt
]{commitbar}

\crefname{section}{\S}{\S\S}
\Crefname{section}{\S}{\S\S}
  
\begin{document}

\maketitle

% \textcolor{red}{Still need to update our definition of curiosity with an update of the tradition curiosity from RL to LLM Agents in both closed-tasks and open-ended environments.}

% ========================d=============================================

\begin{abstract}
LLM agents are increasingly used for open-ended tasks, very common in research scenarios where they are required to explore, investigate, or act curiously. At inference time, an LLM is a fixed map from context to next-token distribution, the pretrained weights determine the function, and only its conditioning changes as context accumulates. Mechanisms like temperature, best-of-N, tree search, or RAG vary how the distribution is sampled or what it is conditioned on, but do not update the agent's model of the environment. We argue that current LLM agents do not explore curiously; they explore stochastically. From how curiosity is understood in the RL literature, it requires a predictive structure over how the environment responds to the agent's actions, which improves online from current interaction, drives action selection through its predicted improvement, and persists beyond the context window. Additionally, in its open-ended form, the agent should intrinsically originate what to investigate. Thus, we propose a framework defining curiosity in LLM agents, reserving the term for systems that satisfy it. In doing so, we identify the research directions its implementation opens up. 
\end{abstract}

\begin{figure}[h]
\vspace{-10pt}
    \centering
    \includegraphics[width=0.84\linewidth]{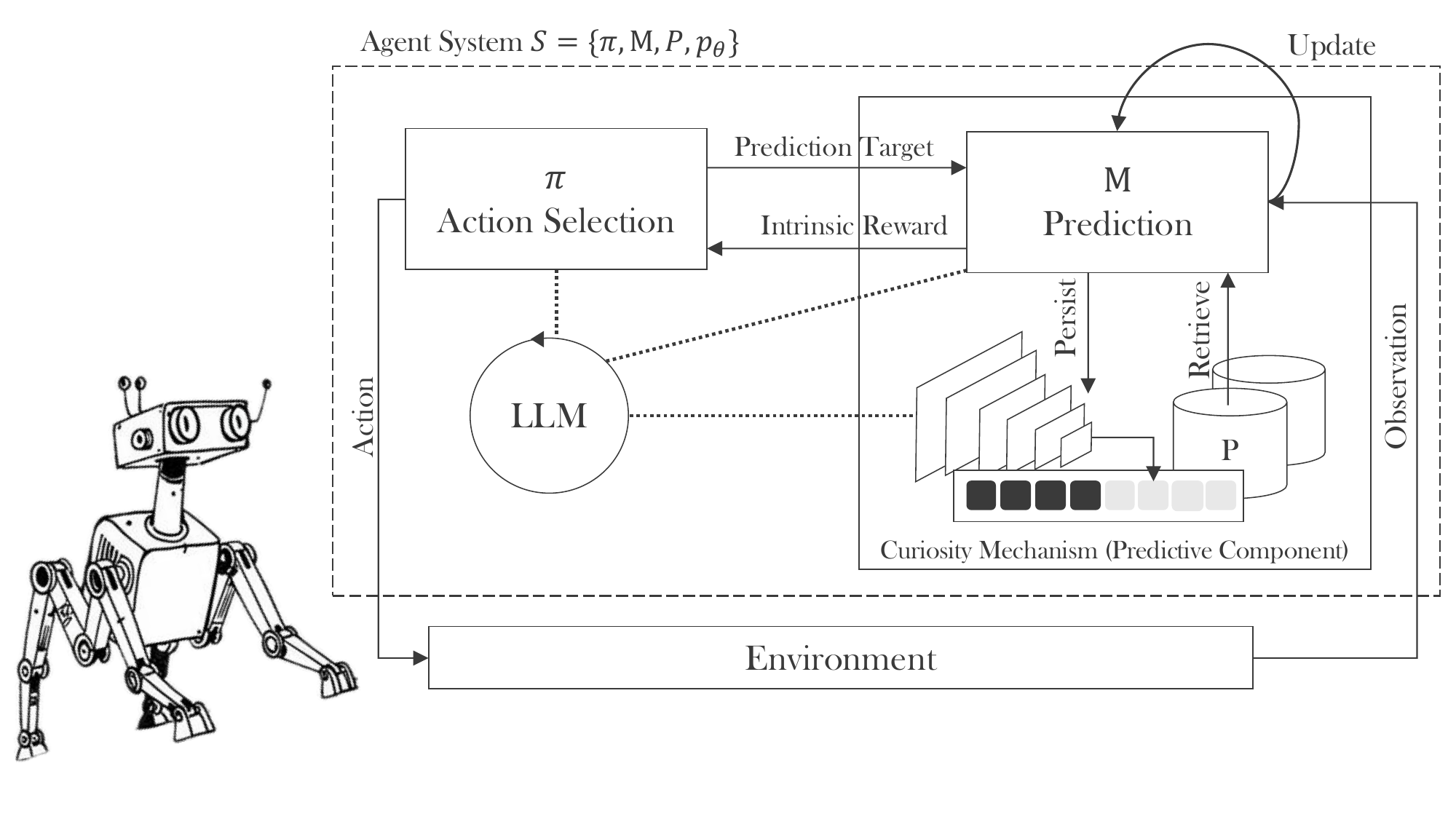}
    \caption{Design of conceptual components for a curious agent system $\mathcal{S} = \langle p_\theta, \mathcal{M}, \mathcal{P}, \pi \rangle$. The predictive component $\mathcal{M}$ targets environment responses [C1] and updates from observations [C2]; its predicted improvement supplies the intrinsic reward shaping action selection through $\pi$ [C3]; persistent state $P$ retains improvements beyond the context window [C4]. In open-ended settings, it can also design its own prediction targets and environments [C5].
    }
    \label{fig:placeholder}
\vspace{-10pt}
\end{figure}

\section{Introduction}
\label{sec:introduction}
% Three jobs: motivate, state position, preview structure. Open with the open-endedness shift — research assistance, scientific discovery, autonomous investigation — and the vocabulary problem that follows. State the position in bold ("LLM agents do not explore curiously; they explore stochastically. Curiosity, in the sense the term inherits from the RL tradition, requires four architectural commitments…"). End with a short roadmap of the rest.

% Paragraph 1: Motivation
% \textit{LM agents are moving from closed tasks to open-ended ones
% }

As LLM agents are deployed in more scenarios, they are increasingly requested to have agency and autonomy. However, the agents are not designed with that scenario in mind. In particular, as they go into more open-ended domains, such as research, we expect AI agents to propose and validate hypotheses, conduct experiments, and analyze results for further inquiry. Recent systems such as The AI Scientist \cite{lu2024ai}, CORAL \cite{qu2026coral}, or AlphaEvolve \cite{novikov2025alphaevolve} frame their goal as open-ended exploration and report results that suggest the field is moving beyond task completion as the primary capability of interest. We identify a clear trajectory going forward, moving from requesting agents to complete tasks to open-ended tasks where the agent has to figure out how the successful task is defined. This shift in capability is not about better task-following, but instead the agent needs to decide what to do.

% Paragraph 2: Vocabulary issue (symptom)
% \textit{the field's vocabulary of curiosity and exploration applied to LLM agents 
% }

% With the evolution of capabilities, the field has responded with the use of specific vocabulary for \textit{curiosity, exploration, novelty} or \textit{intrinsic motivation}. Methods like Tree-of-Thoughts \cite{yao2023tree} or Graph-of-Thoughts \cite{besta2024graph} are framed as exploration of reasoning paths; retrieval-augmented memory is offered as a substrate for learning from experience  \cite{lewis2020retrieval}, or intrinsic-reward post-training methods are explicitly under the name of curiosity-driven \cite{dai2025cde}. The vocabulary is inspired and borrowed from the reinforcement-learning tradition and the previous works on artificial curiosity \cite{schmidhuber2010formal}. The borrowing has happened quickly and with little examination of whether the architectural commitments that originated the words are met. \textcolor{red}{Not in love with this paragraph. I think we should change this paragraph into the distinction of interaction vs. adaptation}

With this evolution of capabilities, the field has responded by adopting a vocabulary of \textit{curiosity, exploration, novelty}, and \textit{intrinsic motivation}. Search-style methods like Tree-of-Thoughts \cite{yao2023tree} or Graph-of-Thoughts \cite{besta2024graph} are framed as exploration of reasoning paths; Retrieval-Augmented (RAG) memory is offered as a  foundation for learning from experience  \cite{lewis2020retrieval}; intrinsic-reward post-training methods are explicitly labeled curiosity-driven \cite{dai2025cde}. These terms are inspired by and borrowed from the reinforcement-learning tradition and the previous works on artificial curiosity \cite{schmidhuber2010formal}, but the architectural commitments that gave them their meaning have not been carried over.
% Nonetheless, we identify that some of the architectural commitments that created these terms have not been transported accordingly and have development potential for LLM agents.

% Paragraph 3: Distinction in science A vs B. Interaction is not adaption, conditioning is not learning.
Traditionally, scientific research can be divided into two distinct tasks. On one side, (I) \emph{finding answers to given questions}, and on the other side (II) \emph{coming up with good questions}. Current LLM agent development is focused on improving their performance on (I), but it is not so focused on fulfilling (II). At inference time, an LLM samples from a fixed distribution determined by its pre-trained weights and conditioned on the given context. In agentic settings, the agent interacts with its environment; it takes actions, receives observations and conditions subsequent generations on what it has observed. Exploration in this sense happens during interaction. But it happens at the level of which inputs enter the predictive computation, not at the level of the computation itself. The function from context to next-token distribution remains unchanged. We call this distinction \emph{interaction-level exploration} versus \emph{model-level adaptation}. This distinction reflects an architectural fact about LLMs, that they discover useful information in an environment but observe without learning \cite{englander2026agents}. What varies in these mechanisms is how the model is sampled, not what the model encodes about the environment. 
% Discovery without adaptation, observation without learning.

% Paragraph 4: position in bold
% \textit{ state the position formally.
% }
In this paper, we argue that current LLM agents do not explore curiously, they explore stochastically. Curiosity, in the sense inherited from the RL tradition, requires that the agent maintain a model of the domain it investigates, improve it from online interaction, drive action selection through its predicted improvement, and retain the knowledge in some way across the episode. In its open-ended form, it requires a further commitment that the agent originate what to investigate without being externally specified. The field should reserve the term \textit{curiosity} for systems that satisfy these commitments to some degree, and recognize each of these commitments as an open problem for the future development of LLM agents.

% In the following pages of this paper, we first adapt the curiosity-as-predictive framework from the RL tradition into the LLM-agent setting, deriving the commitments for developing \emph{curious agents}. We then state the architectural fact about inference-time computation that distinguishes interaction-level exploration from model-level adaptation, and use it to evaluate the current approaches in their fulfillment of the defined commitments, identifying how closely existing work is engaging them. Third, we examine what satisfaction looks like in existing partial vindications and engage alternative views. Finally, we recognize the potential impact of this research direction, highlighted by the frameworks, and propose possible future work.

\begin{mybox}{Paper Position}
Current LLM agents do not explore curiously. They explore stochastically, changing how they sample from a fixed distribution instead of updating a predictive model based on interaction. The field's curiosity vocabulary should be reserved for systems that keep a model of the domain under investigation. These systems improve it from online interactions, use it to drive action selection, and retain it throughout the episode. In its open-ended form, curiosity further requires that the agent generate the prediction targets, instead of receiving them externally.
\end{mybox}

In this paper we do not introduce a new mechanism, as each ingredient of the framework exists in the curiosity tradition. Instead, we adapt it to LLM agents, whose conceptual environments, open-ended action spaces, rich priors, and undetermined prediction targets differ from the settings they were developed for. From these differences, we derive commitments that make the claims of curiosity explicit. The remainder of this paper discusses the requirements of curiosity from the RL tradition and its adaption to an LLM-agent setting (\cref{sec:curiosity-requires}), deriving five commitments for \emph{curious agents} (\cref{sec:curiosity_framework}). We then use these commitments to diagnose the current practice (\cref{sec:current_practice}), and discuss alternative views (\cref{sec:alternative_views}). We close with the open problems and future research directions (\cref{sec:research_directions}).

\section{What Curiosity requires}
\label{sec:curiosity-requires}

The vocabulary of \emph{curiosity} and \emph{exploration} comes from decades of research in reinforcement learning, intrinsic motivation, and robotics that originated these terms. This section identifies the traditional requirements of curiosity and adapts them for LLM-based agents.

\subsection{Curiosity at the System Level}
\label{sec:curiosity-at-system-level}

An LLM at inference time is a function from context to next-token distribution $p_\theta(x_t \mid x_{<t})$, where $\theta$ are the pretrained weights, $x_{<t}$ is the current context, and $p_\theta$ is the predictive distribution over the next token. Mechanisms often described as performing \textit{exploration} in the reasoning space like temperature sampling, tree search \cite{yao2023tree}, or retrieval augmentation \cite{lewis2020retrieval}, change how $p_\theta$ is sampled or what it is conditioned on, producing action distributions of the form $\pi(a_t \mid x_{<t}; \phi) = f\!\left(p_\theta(\cdot \mid x_{<t}),\, \phi\right)$, where $\phi$ is a sampling configuration and $f$ is a transformation (sampling rule, search strategy, retrieval mechanism) that produces an action distribution from $p_\theta$'s output. In multi-step scenarios, the context $x_{<t}$ grows across the trajectory as actions are taken and come back from the environment, incrementally conditioned on the accumulating history. This is genuine interaction with an environment, in the traditional sense of environment \emph{exploration}. However, in all cases, $\theta$ is invariant, and the function $p_\theta$ remains the same function before and after the agent observes any response from the environment. What varies in these cases is $\phi$ and the context $x_{<t}$ on which $p_\theta$ is conditioned.

We refer to this distinction as \emph{interaction-level exploration}, which contrasts with the \emph{model-level adaptation}. Interaction-level exploration is the variation in actions, contexts, sampling configuration, or trajectories during the agent's behavior, while maintaining the predictive function fixed. Model-level adaptation is the update of the predictive structure within the system such that the predictions about the environment improve over the episode. Throughout this work, we develop on top of model-level adaptation, under the premise that interaction-level exploration is necessary but not sufficient for curiosity given the traditional understanding of it. Whether some in-context computation could nonetheless implement some aspects of model-level adaptation is an open question that we comment on in Section \ref{sec:alternative_views}.

\begin{callout}
    \vspace{10pt}
    \textbf{Interaction-level Exploration} vs. \textbf{Model-level Adaptation}. Interaction-level exploration varies actions and contexts while the predictive function remains fixed. Model-level adaptation updates the predictive structure itself from environment interaction.
    \vspace{10pt}
\end{callout}

Thus, when building LLM agents, we treat the LLM as a component of a larger system that can be represented as: 

\begin{equation}
    \mathcal{S} = \langle p_\theta,\, \mathcal{M},\, \mathcal{P},\, \pi \rangle,
    \label{eq:system}
\end{equation}

where the LLM, $p_\theta$, forms part of the system, alongside a predictive structure $\mathcal{M}$ over the environment responses, persistent state $\mathcal{P}$, and an action-selection policy $\pi$. We treat the system $\mathcal{S}$, irrespective of the methods used to implement each one of the components, which may or may not contain different calls to the LLM. For instance, the predictive structure $\mathcal{M}$ can live in many places, like in natural-language hypotheses updated by Bayesian inference \cite{levy2025worldllm}, in distilled-memory representations, structured data, latent variables, or in different updates of $\theta$. 

\subsection{The Curiosity Tradition}

Curiosity can be defined as the intrinsic drive to explore, seek understanding, and ask questions in order to learn and make sense of the unknown. In psychology, \citet{berlyne1960conflict} described curiosity as a motivational drive triggered by collative variables such as novelty, surprise, complexity, and incongruity that create a state of epistemic conflict or uncertainty, which increases the level of \emph{arousal}. People search for an optimal arousal level such that when it is too high, they seek information to reduce uncertainty (\emph{curiosity}), or when it is too low, they search for how to increase it (\emph{exploration}). 

For decades, researchers have been exploring how to transform curiosity into a computational feature \citep{schmidhuber2020curiosity_blog}. This translation is not trivial, but there is a notable convergence on what curiosity as an intrinsic reward requires. In early formulations, the agent is rewarded for encountering surprising observations \cite{schmidhuber1991possibility}, grounding the intrinsic motivation in the prediction error of a learnable world model. This formulation faces a well-known difficulty in stochastic environments, known as the \emph{noisy TV problem} \cite{schmidhuber2010formal, burda2018large}. The reward can cause the agent to focus on irreducible noise because its prediction error remains high. This was later refined by introducing the \emph{learning progress} which does not reward the error itself, but its rate reduction \cite{schmidhuber1991curious}. This shift relocated curiosity from ``what surprises the agent'' to ``what the agent is coming to understand''. Subsequent works have formalized this in various ways, such as expected information gain about the environmental dynamics \cite{storck1995reinforcement,itti2005bayesian, sun2011planning} or developmental skill acquisition \cite{oudeyer2007intrinsic}. But the core structural claim is about getting better at predicting the environment from interacting with it.

\citet{schmidhuber2010formal} provides the formalization of the framework, unifying the threads of compression progress where the agent maintains a predictive component of the environment, and intrinsic reward proportional to the rate at which the agent's history shortens as new observations arrive. Additionally, PowerPlay \cite{schmidhuber2013powerplay} extends the framework further to the case when the agent must originate its own environments, generating its own questions in pursuit of compression progress. We see a similar behavior in \citep{herrmann2023learning} proposing an ``artificial scientist'' that does not only answer given questions, but continually invents its own questions driven by intrinsic rewards for surprising outcomes.

In these formulations from the traditional understanding of computational curiosity, we identify four recurring ingredients that are defining features: (i) a learnable predictive component over the environment, (ii) online update of that component from current interaction, (iii) intrinsic reward derived from the system improvement, (iv) action selection in relation to the reward. An extra fifth element (v), the generation of prediction targets, is required when the environment does not provide them. These are the core elements identified from the curiosity tradition. Proposals for learned world models share the same predictive core, using predictions of the environment to plan \cite{ha2018world}, but our focus is on what is required for that prediction to drive exploration in LLM agents.

\subsection{Adapting the Framework to LLM Agents}
\label{sec:curiosity-adaptation}

The artificial curiosity tradition was developed for problems of embodied agents that interact in environments with regular state-transition dynamics. However, the community is currently applying these ideas to LLMs, and we need make the differences explicit to inform our adaptation:

\begin{itemize}[leftmargin=*,itemsep=.3em]
    \item \textit{State-space environments}. The original predictive components were designed for pixel transitions, sensor readings, or game states. LLM agents operate in more conceptual environments such as codebases, literature, web ecosystems, or scientific domains. These environments have some notion of dynamics, but they are mainly conceptual and rarely admit tidy state-space representations.
    \item \textit{Given action spaces}. Embodied agents have a discrete set of actions, and the agent selects from a well-defined set. By contrast, LLM agents act through natural language, tool calls, and queries. This expands the action space, and its structure is something the agent must reason about. 
    \item \textit{No priors}. Embodied-agent formulations typically assume that the agent starts ignorant and learns from scratch. LLMs, however, have a vast amount of general-knowledge data pre-loaded in their weights, which is already doing real work before any episode begins.
    \item \textit{Externally given prediction targets}. In embodied environments, the prediction target is provided by the environment itself. State transitions are what the agents are meant to learn and model. In an open-ended LLM-agent setting, the prediction target is itself undetermined; the agent must decide what subgoals, hypotheses, or mechanisms are worth understanding. 
\end{itemize}

These differences motivate three adaptations of the framework, each grounding one of the commitments derived below:

\begin{enumerate}[leftmargin=*, label=(\Alph*)]
    \item \textit{Prediction target}. The prediction does not need to target specific state transitions or operate in a fixed action space. Now, the prediction target can be a model of how a codebase responds to commands, a hypothesis about a physical system, or any other structure whose accuracy improves as the agent gathers evidence. What survives the generalization is that the agent must predict how the environment responds to the agent's actions, not properties of the agent's own outputs, which rules out signals defined over the LLM's own generation.
    
    \item \textit{Priors}. Pretraining gives the agent a head start, but this cannot be confused with episode-time learning. The agent now has to distinguish what the \emph{agent already knows}, from what \emph{the agent learned from interacting with the environment}. Without this distinction, improvement due to priors could be mistaken for learning. 
    
    \item \textit{Open-endedness}. In embodied environments, the environment determines what is worth investigating. In open-ended settings, nothing plays this role,  so the agent must choose its own lines of inquiry, and that choice becomes part of curiosity and not something external to it.
    
\end{enumerate}

\section{The Curiosity Framework: Commitments for Curious LLM Agents
}
\label{sec:curiosity_framework}

From the curiosity tradition and our adaptation, we derive five architectural commitments for curiosity in LLM agents. The first four are derived from the ideas on prediction improvement driving actions. The fifth one derives from the open-endedness adaptation and applies specifically to settings where the prediction targets are not externally supplied. 
%The commitments are computational, rather than architectural; they specify what the system should compute and not how it should be done or where the computation must be executed. 

\begin{commitbar}
\vspace{0.2em}
\textbf{[C1] Predict Environment}. The agent maintains a predictive component over how the environment responds to its actions, with the prediction target being environmental dynamics rather than properties of the agent's own generation.
\end{commitbar}

The agent maintains predictions about how the environment responds to its actions. The predictive structure can live anywhere in the system, not necessarily as part of the LLM. It can live in parametric updates, hypothesis state, a memory, or structured data. But the target of the prediction is the environmental dynamics, not the agent's own outputs. A system can be uncertain about its own outputs in ways unrelated to what it knows about the world it is acting in.

\begin{commitbar}
\vspace{0.2em}
\textbf{[C2] Update from Interaction}. The prediction improves within the episode, and the improvement is caused by current observations.
\end{commitbar}

Predictions improve during the episode, grounded in the current interaction with this environment and not retrieved from priors. By online, we mean within-episode, and grounded means caused by the observations of the environment. This rules out improvements that derive from pretraining alone, as the LLM already encodes a substantial amount of world knowledge in its weights. It also rules out improvements that come from retrieval over a fixed external corpus. Both can produce better-conditioned predictions without any update driven by what the agent has actually observed. The agent has to learn something from the current interaction.

\begin{commitbar}
\vspace{0.2em}
\textbf{[C3] Act to Learn}. Action selection is shaped by where the predictive component is expected to improve, so the agent acts to learn rather than learning while acting.
\end{commitbar}

The first two commitments give the agent a predictive component and a mechanism to update it. This commitment closes the loop, stating that actions are chosen because their outcomes are expected to be informative for the predictive component, and the resulting observations feed back into the update. Without this, updates may occur, but they do not shape what the agent does next. This separates the framework from systems where action selection is driven entirely by task reward. Curiosity requires that action selection respond to expected predictive improvement, whether measured as prediction error, information gain, or learning progress.

\begin{commitbar}
\vspace{0.2em}
\textbf{[C4] Retain the Improvement}. Improvements to the predictive component persist in some form beyond the context window in which they were formed.
\end{commitbar}

The persistence requirement is about availability for future action selection, not about storage. Improvements derived from interaction remain accessible to shape what the agent does later, and this access cannot rely on the observations remaining in context. The commitment is method-agnostic, but it rules out systems in which the only source of improvement is the immediate context. The commitment requires the system to carry the improvement beyond the boundary of the context.

\begin{commitbar}
\textbf{[C5] Choose What to Investigate} (\textit{Open-Endedness}). The agent originates the prediction target itself, instead of receiving it from the environment or from a human specification of the problem.
\end{commitbar}

The first four commitments together define curiosity in its task-driven form, where an agent investigates a given domain,  improving its predictions about that domain, and acting to drive that improvement. In the division of the two important scientific tasks, this would correspond to (I) finding answers, which is externally specified. The fifth commitment addresses the second task, (II) coming up with good questions. It applies specifically to the open-ended setting where prediction targets are not externally supplied. The agents originates what to model and what counts as worth understanding. It can propose its own fitness function, benchmark, or problem domain. This commitment does not raise the bar for curiosity in general, an agent that designs experiments and hypotheses within a human-provided domain is a curious agent in the task-driven form under C1–C4. C5 distinguishes an additional frontier of open-ended curiosity, where the prediction target itself is not given.

% \paragraph{[C5] Choose what to investigate (open-endedness)} The agent originates what to predict without an externally specified task. Action selection is not only defined by improvement-in-prediction within a given domain; the agent itself selects what to model, what counts as worth understanding. This commitment is what distinguishes open-ended curiosity from task-driven exploration. 

\paragraph{$\blacklozenge$\ Curious Agents} We call systems satisfying C1–C4 \emph{curious agents in the task-driven form}, and systems satisfying C1–C5 \emph{open-ended curious agents}, with C5 extending curiosity to settings where the prediction target is not supplied. The category is computational, not behavioral; a system is a curious agent if it demonstrably satisfies the commitments, and the rest of the paper takes the commitments as our criterion for the term.

\paragraph{Example} To make this concrete, consider an agent investigating a new material:
\begin{itemize}[leftmargin=*, itemsep=2pt, topsep=3pt]
    \item \textbf{Stochastic Exploration.} The agent generates diverse candidate hypotheses (e.g., via temperature or best-of-N), tests them in arbitrary order, and appends failed experiments to its context. It has no prediction of which experiment would most reduce its uncertainty.
    \item \textbf{Curious Exploration.} The agent maintains competing hypotheses about the material [C1], predicts the outcome of candidate experiments under each and runs the most informative one [C3], updates its hypotheses on the result [C2], and carries the update forward to shape the next experiment [C4].
\end{itemize}
Both agents produce multiple actions, interact with the environment, and may use memory, but only the second closes the loop from prediction to action update, selecting actions that maximize learning.

\section{Diagnosing Current Practice}
\label{sec:current_practice}

% The vocabulary of curiosity, exploration, and intrinsic motivation is currently being applied to LLM-agent work that varies enormously in what it actually computes. This section maps that variation against the five commitments developed in §3, identifying where current approaches engage the framework and where they fail it. The goal is not to dismiss any tier but to make architecturally distinct things architecturally distinct, so that the question what kind of curiosity does this system implement? has a precise answer.

% \paragraph{Tier 1: Sampling as curiosity} 

% Tier 2: Storage as curiosity

% Tier 3: Trying 

% Tier 4: Open-endedd Research

% Tier 5: Partial vindications

In the previous sections, we went over what is required for curiosity and the translation of the framework to LLM agents. We now turn that framework on the field to see how current works relate to our commitments. Applying the commitments requires distinguishing two kinds of systems:

\begin{itemize}[leftmargin=*, itemsep=2pt, topsep=3pt]
    \item \textbf{Explicit systems.} The components of the system, $\mathcal{S}$, the predictive component $\mathcal{M}$, update rule, and the policy $\pi$ are externally inspectable. Whether the commitments are met can be analyzed from the system's design.
    
    \item \textbf{Implicit systems.} The relevant computation may happen inside the architecture, and causal identification is substantially harder. In these cases, the conclusion is that curiosity \emph{has not been demonstrated}, not that the system is not curious.
\end{itemize}

The systems we discuss here are explicit, and our verdicts follow from the designs stated in their own papers. We close with examples that show how to achieve the architectural targets.

\paragraph{The wrong prediction target} A common pattern is the implementation of methods using \emph{curiosity} or \emph{exploration} signals that are computed over the model's own output rather than over the environmental responses from the actions. Curiosity-Driven Exploration (CDE) \cite{dai2025cde} uses per-response perplexity as a curiosity bonus, computing novelty over generated trajectories; CoVo \cite{zhang2025consistent} combines a consistency-of-trajectory intrinsic reward with a perplexity-based curiosity bonus, both computed over the model's own reasoning paths; and a study of ideation diversity measures the variation of model proposals \cite{audran2025does}. In these cases, the signal driving exploration is internal to the language model and mainly tracks whether the model produces varied output, and not whether the agent is uncovering new knowledge in the environment.

This differs from our framework, where commitment [C1] concerns a prediction over environmental responses, distinct from the language model's generation distribution, improving through interaction. A coverage objective over output space gives the agent no insight into how the world responds, because the question is never asked. In these cases, the systems are doing real work, like improving robustness, escaping local minima, or broadening hypothesis search, but the work is happening at a different level than curiosity understood in the traditional sense. 

One could argue that diverse generation is doing the productive work by broadening the search space. \citet{audran2025does} documents this by mentioning that ideation diversity provides robustness through redundancy. This is indeed useful. But we would say this differs from what is known as computational curiosity. This idea of robustness through redundancy works by sampling more broadly from a fixed distribution, which can hardly change the agent's model of the world.

\paragraph{Memory mistaken for adaptation} Another pattern identified is the use of persistent storage as the basis of online learning. A popular standard is to maintain a memory of reflections on prior failures \cite{shinn2023reflexion}. Similar works like Voyager \cite{wang2023voyager} accumulate a growing library of skills of executable code, retrieved by task-embedding similarity. These systems retain the structure of the episode from multiple attempts, and their stored content updates during interaction. 

The architectural question is what gets updated. The framework describes a predictive component that improves from the interaction with the environment, with the improvement shaping the next actions. In this line of work, we see that the retrieved content can change what the policy does, but the loop from prediction to action, which involves selecting actions because their outcomes are expected to reduce predictive error, is not present. Neither Reflexion nor Voyager contains a component that scores predictions about environmental responses, so there is no signal to which action selection could respond. Action selection is driven by task reward and the augmented context, not by where predictions are expected to improve.

Adjusting which stored traces to retrieve is a meaningful operation, but it is a different operation from selecting actions whose outcomes the agent expects to be informative. The first changes how the agent acts on what it already believes, the second changes what the agent investigates next. The limitation is therefore not memory itself, but what is stored and how it is used. A memory-based system could satisfy the commitments, but it is eventually limited because memory acts only through the bounded context, so at scale, weight updates may be needed for learning to last. The relation between memory and action-driven prediction remains an open architectural question.

\paragraph{Externally-supplied inquiry direction} A third direction is the range of systems that orchestrate exploration over a problem space defined by humans, which have made a significant impact and outstanding progress. The AI Scientist \cite{lu2024ai}, its v2 successor \cite{yamada2025ai}, AlphaEvolve \cite{novikov2025alphaevolve}, FunSearch \cite{romera2024mathematical}, or Coral \cite{qu2026coral}, to name some popular ones, all execute impressive long-horizon work. They generate research hypotheses, evolve programs toward score-maximizing objectives, and discover mathematical algorithms inside human-defined search spaces. 

Several of these systems explicitly position themselves as open-ended discovery, targeting  open scientific problems. This sense of open-endedness, where there is a large search space, unbounded progress against an unknown optimum point, is real, and these systems make great progress. They can generate their own subgoals within a task, which candidate to inspect or what is still to distill. This is a genuine inquiry direction at the sub-goal level. Under our open-ended commitment [C5], the agents should eventually be able to formulate the problems themselves, for instance by proposing their own fitness function, benchmark, or problem domain, and refine them based on their own predictive structure. In current systems, the frame is still supplied by humans and remains fixed; the agent's autonomy operates inside it. The framework's open-ended form asks for agents that generate inquiry direction, achieving the scientific frontier of generating their own unsolved questions.

\paragraph{What architectural engagement looks like} The framework's commitments describe an achievable target, not an aspirational one. WorldLLM \cite{levy2025worldllm} shows an example of the four task-driven commitments, without modifying the weights. The system maintains a set of natural-language hypotheses over environmental responses, which are updated during interaction through an LLM inference loop, conditioned on observed outcomes. With no extrinsic reward, action selection is driven entirely by intrinsic signals from these hypotheses, where prediction error performing the best, similar to \citet{schmidhuber1991possibility}, while a learning-progress variant underperformed for estimation reasons. SENSEI \cite{sancaktar25a} shows the potential with parameter updates, training a world model learned from interaction, with an exploration policy that optimizes high epistemic uncertainty, so improvements persist in the weights by construction. AutoDiscovery \cite{agarwal2026autodiscovery} applies the loop to data-driven discovery. The LLM's belief in a hypothesis is generated before and after an experiment runs on real data, and the resulting Bayesian surprise rewards a tree search over hypotheses, with results persisting in the tree and not in an updated model. In all three, what the agent does next depends on what it expects to learn.

%WorldLLM \cite{levy2025worldllm} provides a concrete demonstration that a system can engage with the four task-driven commitments without modifying the language model's weights. It does so in small text-based environments, and whether the approach scales remains open. The system maintains a set of natural-language hypotheses about its environment, which become an external structured predictive component over environmental responses, distinct from the language model's generation distribution. The hypotheses update during interaction, conditioned on observed outcomes, through an LLM-driven inference loop that proposes revisions when predictions fail. Its environments provide no extrinsic reward, so action selection is driven entirely by intrinsic signals derived from the predictive component. The best performing variant rewards prediction error, similar to \citet{schmidhuber1991possibility}, while a learning-progress variant underperformed for estimation reasons. This closes the action-prediction loop on environmental responses rather than on output novelty. The hypotheses persist across episodes in structured natural-language form, available for further refinement.

None of these fully engages the open-ended commitment [C5].  WorldLLM's hypothesis space is defined by human specification of the environment, and SENSEI's notion of interestingness is fixed before exploration begins. AutoDiscovery comes closest, choosing which questions to pursue without a specified research goal, but within human-supplied datasets. Work on \emph{autotelic agents} \cite{forestier2022intrinsically, zhang2023omni} explores parts of these problems. Nevertheless, engaging  all five commitments simultaneously remains an unsolved task, but the architectural components are available in different forms. Stating the commitments jointly is what makes this gap visible.

\section{Alternative Views}
\label{sec:alternative_views}

\paragraph{In-Context Learning (ICL) is Compression Progress} As an LLM agent accumulates context from observations, prior actions, and retrieved memory, its conditional distribution differs from the one induced by the task input alone, $p_\theta(y \mid x, \text{context})$. The amount of context agents can maintain has grown alongside progress in LLMs, although sometimes at the cost of quality. The conditional becomes more accurate over the course of an episode as more relevant context accumulates. This is, in a precise sense, predictive improvement driven by interaction. It has been shown that in-context learning can be seen as a form of implicit Bayesian inference over latent concepts in the pretraining distribution \cite{xie2021explanation}, where the conditional distribution updates as context accumulates and concentrates on the concept consistent with the prompt examples. Under this view, pretrained knowledge can supply the predictive component [C1], and the in-context update is caused by current observations rather than by priors or retrieval, so it can satisfy [C2] as stated, but with some limitations.

In-context Learning can be seen as an explicit boundary case of the framework, instead of a counterexample. What changes during ICL is the conditioning of the model but not the model itself, the map $p_\theta$ from context to next-token distribution is fixed and the input varies. This is where we can claim that current agents sample from a fixed distribution. Granting [C1] and [C2], the commitments that separate stochastic from curious exploration are [C3] and [C4]. By itself, ICL does not fulfill [C3], the conditional distribution can update as context accumulates, but action selection is not driven by where the predictive distribution is expected to improve. The agent's actions are driven by task reward, by retrieved exemplars, and by the policy's response to the augmented context.  ICL provides the predictive update, but not the action-selection criterion that makes the update operative as curiosity. Second it does not satisfy [C4], the model context limit puts a hard constraint on the persistence of the updates, so improvements that live only in the conditioning are evicted with the observations that produced them. Neither failure can be closed by prompting or sampling differently. The framework does not rule out in-context learning as a direction towards curious agents; on the contrary. A system built around in-context predictive improvement can lead to curiosity when designed around predictive improvement, with action selection shaped by where the improvement is expected and the persistent state. \cite{elmoznino2026can} shows an example toward [C3], where they train a policy to maximize learning progress of a frozen in-context learner, but show this works only when actions do not change the environment, also leaving [C4] unaddressed. 

\paragraph{Tool-using agents are interacting with the environment} Modern deployed agents, such as coding agents, web agents, or scientific agents, have access to tools and interact with an environment. They can launch commands, observe the result, and review their behavior. One could argue that environmental interaction is a basis of curiosity, and these systems already have it. 

Interaction, however, is necessary, but not sufficient. We argue that interaction-level exploration, where an agent performs actions and observes the results, is different from model-level adaptation, where the agent has a predictive component that improves from those observations. Tool usage by itself satisfies the condition of interaction. But a current tool-using LLM agent does not measurably improve a prediction from its observations. This has been reported by \citet{englander2026agents}, highlighting that agents discover information in their environment, but fail to act on it. 

\paragraph{RL-finetuning encodes in weights} At inference time, the map $p_\theta$ is fixed and only its conditioning varies, but RL fine-tuning does update the model weights. Across training, the model's distribution shifts to better match the environment. This can partially match the requirement of prediction improvement from our commitment [C2]. However, there is a mismatch in timescale, as current RL finetuning method still still separates training from deployment. Weight updates from standard RL fine-tuning happen across training episodes, not within an episode in response to current observations. Curiosity is operative in the moment of action selection; the predictive improvement has to be available to shape what the agent does next. Weight updates from yesterday's training run produce a better-calibrated policy, but not a system that revises its predictions in response to what it just observed and acts on the revision.

This drives the argument to approaches closer to Meta-RL and Continual Learning, which complicates this picture in a different way. Meta-RL \cite{schmidhuber1994learning, schmidhuber1997shifting, jiang2025meta, laskin2022context} learn to implement an RL algorithm at inference time, allowing easy deployment to new tasks. Continual learning \cite{ring1994continual, ring1997child, kirkpatrick2017overcoming,jang2021towards} targets a sequential task stream and seeks weight updates that persist throughout different tasks without catastrophic forgetting. Streaming online RL pushes weight updates to episode time directly. Where standard RL fine-tuning places the predictive update across training episodes, these approaches push the update closer to the episode time.

% \paragraph{Other froms of curiosity} \textcolor{red}{unsure about writing down this one}

\section{Research Directions} 
\label{sec:research_directions}
\vspace{-5pt}

Throughout this paper, we have discussed what curiosity in LLM-based agents requires. We have kept the formulation technology-agnostic and used it to diagnose where current practice falls short. The framework names four commitments and one extension. We identify some areas of research derived from these commitments.

\subsection{Open Problems}
\label{sec:open_problems}

\paragraph{Recognizing predictive components} \textit{Would we recognize a curious system if we saw one?} \newline
We have established a framework of curiosity for LLMs that is agnostic to where the predictive component lives, whether it is in parametric updates, hypothesis state, or structured memory. As a result, how we recognize it depends on the system. For \emph{explicit} systems, the commitments can be checked against the design, as we do in \cref{sec:current_practice}. For \emph{implicit} systems, where the  computation may happen inside the architecture, causal identification is substantially harder. In these cases, we suggest behavioral tests, which apply without inspecting internals. They are a starting point, where passing them shows that a system behaves as the commitments require, but not where or how its predictive component is implemented. Until an implicit system is tested, the correct conclusion is that curiosity has not been demonstrated, not that the system is not curious. In particular, if a transformer-based system contains a predictive component, how could we evaluate it? Mechanistic interpretability \cite{olsson2022context} can provide tools to better evaluate LLMs in this context, but developing such operational notion remains an open question.

% preamble: \usepackage{tcolorbox}  \tcbuselibrary{raster}
\begin{tcolorbox}[colback=white, colframe=gray!50, colbacktitle=gray!25,
                  coltitle=black, fonttitle=\small\bfseries,
                  boxrule=0.5pt, sharp corners,
                  left=6pt, right=6pt, top=6pt, bottom=5pt,
                  toptitle=3pt, bottomtitle=3pt,
                  title={Behavioral tests for the task-driven commitments}]
\begin{tcbraster}[raster columns=2, raster equal height=rows,
                  raster column skip=6pt, raster row skip=6pt,
                  colback=gray!4, colframe=gray!35, colbacktitle=gray!12,
                  coltitle=black, fonttitle=\small\bfseries,
                  boxrule=0.3pt, sharp corners,
                  left=5pt, right=5pt, top=3pt, bottom=3pt,
                  toptitle=2pt, bottomtitle=2pt]
\vspace{5pt}
\begin{tcolorbox}[title={[C1] Predict Environment}]
\small Before acting, the agent predicts the outcome of action $a$ in state $s$; the prediction is scored against the observed response.
\end{tcolorbox}
\begin{tcolorbox}[title={[C2] Update from Interaction}]
\small Prediction error on held-out responses drops after a controlled set of observations, more than for a control agent that does not receive them.
\end{tcolorbox}
\begin{tcolorbox}[title={[C3] Act to Learn}]
\small Among actions with similar task value, the agent prefers the more informative one, and the preference disappears when the intrinsic signal is ablated.
\end{tcolorbox}
\begin{tcolorbox}[title={[C4] Retain the Improvement}]
\small Prediction accuracy survives a context reset that keeps the persistent state $\mathcal{P}$, but not one that also wipes $\mathcal{P}$.
\end{tcolorbox}
\end{tcbraster}
\vspace{5pt}
\end{tcolorbox}

\paragraph{Within-episode measurement} \textit{Does the agent's prediction actually improve as it acts?}  \newline
In large part, the commitments presented in the framework assume that the predictions improve from current observations [C2] and that action selection responds to where the improvement is expected [C3]. Current agent benchmarks for agentic discovery and exploration, like DiscoveryWorld \cite{jansen2024discoveryworld} or MLAgentBench \cite{huang2023mlagentbench}, do not expose these dynamics. They are mainly conceived to score their outcomes, not predictive trajectories. Even ambitious open-ended works like the AI Scientist \cite{lu2024ai} are scored on the artifacts produced, not what the agent discovered. The key difficulty is attribution, distinguishing improvement caused by the current observations from improvement attributable to priors, retrieval, or in-context patterns. 

\paragraph{Combining with task reward} \textit{How does the curiosity signal coexist with the agent's actual job?}  \newline
Real agents have task rewards, user requests, and score-maximization objectives (extrinsic signal). The intrinsic signal has to deal with the extrinsic rewards without overwhelming or being overwhelmed by them. This is a version of the exploration vs. exploitation trade-off. The RL literature offers partial answers. Some forms of adversarial curiosity, in which two modules bet against each other on the outcomes of computational experiments, were even shown to accelerate the intake of external reward \cite{schmidhuber2003exploring}. Meta-RL and continual-learning approaches bring updates closer to episode time. However, a key problem when applying intrinsic rewards to LLMs is how an intrinsic-improvement signal is weighted against an extrinsic objective in agents whose action space is open natural language and tool calls. This area requires further ideas for deployment.

\paragraph{Choosing what to predict} \textit{What should the agent be trying to predict in the first place?}  \newline
A key aspect of curiosity is that it requires improvement over an environment's predictions [C1], but the target of those predictions remains unspecified; natural-language hypotheses, latent variables, or a distilled memory can qualify, but we are not limited by them. For instance, WorldLLM \cite{levy2025worldllm} gives the agent structured natural-language hypotheses to refine, but the hypothesis space comes from a human specification of the environment. It remains an open problem to select what the agent should predict, and even more in the open-ended scenario [C5]. Beyond single agents, curious agents could also learn together, sharing what they learn about the environment so that each one’s progress informs what the others investigate, as explored for cooperative bandits with heterogeneous agents \cite{pmlr-v119-dubey20b}.

\subsection{The path of the field forward}

Few LLM agents today can satisfy these commitments in a meaningful sense (\cref{sec:open_problems}). But agents are deployed in increasingly open-ended tasks, where the difficulty is no longer to complete a specific task, but to figure out the job and what is worth knowing about it. The gap between sampling and curiosity becomes harder to avoid. However, we do not know which architecture closes that gap. Sampling does not produce what curiosity vocabulary is meant to describe. Reserving \emph{curiosity} costs little, since a claim need only state which commitments it meets, while using the term loosely makes the open problem impossible to state. Building toward systems that satisfy the commitments is worth the work, and worth doing under a name that brings it together.

\section{Conclusion}

Our work argues that current LLM agents do not explore curiously, they explore stochastically. The map from context to output distribution is fixed, while what changes is how it is sampled and what it is conditioned on. From the curiosity tradition, we identify that curiosity requires a prediction over environment responses, updated online from current interaction, driving action selection through the improvement it is expected to yield, and persisting beyond the context that produced it. In open-ended settings, it further requires the agent to produce its own prediction targets. From this, we derived commitments that translate this concept to LLM agents, and used them to diagnose current developments in the field, showing where they fall short or where they approach this concept. This analysis identifies the components of a curious agent, and with them the problems to solve to build more curious agents.

As agents increase their capabilities and are deployed in more scenarios, the challenge shifts from completing specified tasks to deciding what is worth investigating, and from there to closing the gap between task completion and curiosity while keeping that curiosity aligned with human interests.

% [Interesting paragraph] The three patterns documented above are not independent. A system that computes exploration signals over its own outputs has no separate predictive component for online update to operate on; a system that treats memory as adaptation tends also to leave inquiry direction external, because there is no internal predictive structure from which inquiry could originate. The architectural shortfall is coherent: it is what remains when sampling, storage, and orchestration are mistaken for prediction, update, and origination.

\bibliographystyle{plainnat}
\bibliography{references}

@article{lu2024ai,
  title={The ai scientist: Towards fully automated open-ended scientific discovery},
  author={Lu, Chris and Lu, Cong and Lange, Robert Tjarko and Foerster, Jakob and Clune, Jeff and Ha, David},
  journal={arXiv preprint arXiv:2408.06292},
  year={2024}
}

@article{qu2026coral,
  title={CORAL: Towards Autonomous Multi-Agent Evolution for Open-Ended Discovery},
  author={Qu, Ao and Zheng, Han and Zhou, Zijian and Yan, Yihao and Tang, Yihong and Ong, Shao Yong and Hong, Fenglu and Zhou, Kaichen and Jiang, Chonghe and Kong, Minwei and others},
  journal={arXiv preprint arXiv:2604.01658},
  year={2026}
}

@article{novikov2025alphaevolve,
  title={Alphaevolve: A coding agent for scientific and algorithmic discovery},
  author={Novikov, Alexander and V{\~u}, Ng{\^a}n and Eisenberger, Marvin and Dupont, Emilien and Huang, Po-Sen and Wagner, Adam Zsolt and Shirobokov, Sergey and Kozlovskii, Borislav and Ruiz, Francisco JR and Mehrabian, Abbas and others},
  journal={arXiv preprint arXiv:2506.13131},
  year={2025}
}

@article{yao2023tree,
  title={Tree of thoughts: Deliberate problem solving with large language models},
  author={Yao, Shunyu and Yu, Dian and Zhao, Jeffrey and Shafran, Izhak and Griffiths, Tom and Cao, Yuan and Narasimhan, Karthik},
  journal={Advances in neural information processing systems},
  volume={36},
  pages={11809--11822},
  year={2023}
}

@inproceedings{besta2024graph,
  title={Graph of thoughts: Solving elaborate problems with large language models},
  author={Besta, Maciej and Blach, Nils and Kubicek, Ales and Gerstenberger, Robert and Podstawski, Michal and Gianinazzi, Lukas and Gajda, Joanna and Lehmann, Tomasz and Niewiadomski, Hubert and Nyczyk, Piotr and others},
  booktitle={Proceedings of the AAAI conference on artificial intelligence},
  volume={38},
  number={16},
  pages={17682--17690},
  year={2024}
}

@article{lewis2020retrieval,
  title={Retrieval-augmented generation for knowledge-intensive nlp tasks},
  author={Lewis, Patrick and Perez, Ethan and Piktus, Aleksandra and Petroni, Fabio and Karpukhin, Vladimir and Goyal, Naman and K{\"u}ttler, Heinrich and Lewis, Mike and Yih, Wen-tau and Rockt{\"a}schel, Tim and others},
  journal={Advances in neural information processing systems},
  volume={33},
  pages={9459--9474},
  year={2020}
}

@article{dai2025cde,
  title={Cde: Curiosity-driven exploration for efficient reinforcement learning in large language models},
  author={Dai, Runpeng and Song, Linfeng and Liu, Haolin and Liang, Zhenwen and Yu, Dian and Mi, Haitao and Tu, Zhaopeng and Liu, Rui and Zheng, Tong and Zhu, Hongtu and others},
  journal={arXiv preprint arXiv:2509.09675},
  year={2025}
}

@article{schmidhuber2010formal,
  title={Formal theory of creativity, fun, and intrinsic motivation (1990--2010)},
  author={Schmidhuber, J{\"u}rgen},
  journal={IEEE transactions on autonomous mental development},
  volume={2},
  number={3},
  pages={230--247},
  year={2010},
  publisher={Ieee}
}

@article{englander2026agents,
  title={Agents Explore but Agents Ignore: LLMs Lack Environmental Curiosity},
  author={Engl{\"a}nder, Leon and Althammer, Sophia and {\"U}st{\"u}n, Ahmet and Gall{\'e}, Matthias and Sherborne, Tom},
  journal={arXiv preprint arXiv:2604.17609},
  year={2026}
}

@article{berlyne1960conflict,
  title={Conflict, arousal, and curiosity.},
  author={Berlyne, Daniel E},
  year={1960},
  publisher={McGraw-Hill Book Company}
}

@misc{schmidhuber2020curiosity_blog,
  author = {Schmidhuber, Jürgen},
  title = {Artificial Curiosity \& Creativity Since 1990-91},
  howpublished = {\url{https://people.idsia.ch/~juergen/artificial-curiosity-since-1990.html}},
  note = {Accessed: 2026-02-17},
  year = {2021, 2022, 2023, 2025},
  organization = {IDSIA}
}

@article{levy2025worldllm,
  title={WorldLLM: Improving LLMs' world modeling using curiosity-driven theory-making},
  author={Levy, Guillaume and Colas, C{\'e}dric and Oudeyer, Pierre-Yves and Carta, Thomas and Romac, Cl{\'e}ment},
  journal={arXiv preprint arXiv:2506.06725},
  year={2025}
}

@inproceedings{schmidhuber1991possibility,
  title={A possibility for implementing curiosity and boredom in model-building neural controllers},
  author={Schmidhuber, J{\"u}rgen},
  booktitle={Proc. of the international conference on simulation of adaptive behavior: From animals to animats},
  pages={222--227},
  year={1991}
}

@article{burda2018large,
  title={Large-scale study of curiosity-driven learning},
  author={Burda, Yuri and Edwards, Harri and Pathak, Deepak and Storkey, Amos and Darrell, Trevor and Efros, Alexei A},
  journal={arXiv preprint arXiv:1808.04355},
  year={2018}
}

@inproceedings{schmidhuber1991curious,
  title={Curious model-building control systems},
  author={Schmidhuber, J{\"u}rgen},
  booktitle={Proc. international joint conference on neural networks},
  pages={1458--1463},
  year={1991}
}

@inproceedings{storck1995reinforcement,
  title={Reinforcement driven information acquisition in non-deterministic environments},
  author={Storck, Jan and Hochreiter, Sepp and Schmidhuber, J{\"u}rgen and others},
  booktitle={Proceedings of the international conference on artificial neural networks, Paris},
  volume={2},
  pages={159--164},
  year={1995}
}

@article{itti2005bayesian,
  title={Bayesian surprise attracts human attention},
  author={Itti, Laurent and Baldi, Pierre},
  journal={Advances in neural information processing systems},
  volume={18},
  year={2005}
}

@inproceedings{sun2011planning,
  title={Planning to be surprised: Optimal bayesian exploration in dynamic environments},
  author={Sun, Yi and Gomez, Faustino and Schmidhuber, J{\"u}rgen},
  booktitle={International conference on artificial general intelligence},
  pages={41--51},
  year={2011},
  organization={Springer}
}

@article{oudeyer2007intrinsic,
  title={Intrinsic motivation systems for autonomous mental development},
  author={Oudeyer, Pierre-Yves and Kaplan, Frdric and Hafner, Verena V},
  journal={IEEE transactions on evolutionary computation},
  volume={11},
  number={2},
  pages={265--286},
  year={2007},
  publisher={IEEE}
}

@article{schmidhuber2013powerplay,
  title={Powerplay: Training an increasingly general problem solver by continually searching for the simplest still unsolvable problem},
  author={Schmidhuber, J{\"u}rgen},
  journal={Frontiers in psychology},
  volume={4},
  pages={313},
  year={2013},
  publisher={Frontiers Media SA}
}

@article{zhang2025consistent,
  title={Consistent paths lead to truth: Self-rewarding reinforcement learning for llm reasoning},
  author={Zhang, Kongcheng and Yao, Qi and Liu, Shunyu and Wang, Yingjie and Lai, Baisheng and Ye, Jieping and Song, Mingli and Tao, Dacheng},
  journal={arXiv preprint arXiv:2506.08745},
  year={2025}
}

@article{audran2025does,
  title={What Does It Take to Be a Good AI Research Agent? Studying the Role of Ideation Diversity},
  author={Audran-Reiss, Alexis and Armengol-Estap{\~A}{\v{S}}, Jordi and Hambardzumyan, Karen and Budhiraja, Amar and Josifoski, Martin and Toledo, Edan and Hazra, Rishi and Magka, Despoina and Shvartsman, Michael and Pathak, Parth and others},
  journal={arXiv preprint arXiv:2511.15593},
  year={2025}
}

@article{shinn2023reflexion,
  title={Reflexion: Language agents with verbal reinforcement learning},
  author={Shinn, Noah and Cassano, Federico and Gopinath, Ashwin and Narasimhan, Karthik and Yao, Shunyu},
  journal={Advances in neural information processing systems},
  volume={36},
  pages={8634--8652},
  year={2023}
}

@article{wang2023voyager,
  title={Voyager: An open-ended embodied agent with large language models},
  author={Wang, Guanzhi and Xie, Yuqi and Jiang, Yunfan and Mandlekar, Ajay and Xiao, Chaowei and Zhu, Yuke and Fan, Linxi and Anandkumar, Anima},
  journal={arXiv preprint arXiv:2305.16291},
  year={2023}
}

@article{yamada2025ai,
  title={The ai scientist-v2: Workshop-level automated scientific discovery via agentic tree search},
  author={Yamada, Yutaro and Lange, Robert Tjarko and Lu, Cong and Hu, Shengran and Lu, Chris and Foerster, Jakob and Clune, Jeff and Ha, David},
  journal={arXiv preprint arXiv:2504.08066},
  year={2025}
}

@article{romera2024mathematical,
  title={Mathematical discoveries from program search with large language models},
  author={Romera-Paredes, Bernardino and Barekatain, Mohammadamin and Novikov, Alexander and Balog, Matej and Kumar, M Pawan and Dupont, Emilien and Ruiz, Francisco JR and Ellenberg, Jordan S and Wang, Pengming and Fawzi, Omar and others},
  journal={Nature},
  volume={625},
  number={7995},
  pages={468--475},
  year={2024},
  publisher={Nature Publishing Group UK London}
}

@article{forestier2022intrinsically,
  title={Intrinsically motivated goal exploration processes with automatic curriculum learning},
  author={Forestier, S{\'e}bastien and Portelas, R{\'e}my and Mollard, Yoan and Oudeyer, Pierre-Yves},
  journal={Journal of Machine Learning Research},
  volume={23},
  number={152},
  pages={1--41},
  year={2022}
}

@article{zhang2023omni,
  title={Omni: Open-endedness via models of human notions of interestingness},
  author={Zhang, Jenny and Lehman, Joel and Stanley, Kenneth and Clune, Jeff},
  journal={arXiv preprint arXiv:2306.01711},
  year={2023}
}

@article{xie2021explanation,
  title={An explanation of in-context learning as implicit bayesian inference},
  author={Xie, Sang Michael and Raghunathan, Aditi and Liang, Percy and Ma, Tengyu},
  journal={arXiv preprint arXiv:2111.02080},
  year={2021}
}

@article{jiang2025meta,
  title={Meta-RL Induces Exploration in Language Agents},
  author={Jiang, Yulun and Jiang, Liangze and Teney, Damien and Moor, Michael and Brbic, Maria},
  journal={arXiv preprint arXiv:2512.16848},
  year={2025}
}

@article{laskin2022context,
  title={In-context reinforcement learning with algorithm distillation},
  author={Laskin, Michael and Wang, Luyu and Oh, Junhyuk and Parisotto, Emilio and Spencer, Stephen and Steigerwald, Richie and Strouse, DJ and Hansen, Steven and Filos, Angelos and Brooks, Ethan and others},
  journal={arXiv preprint arXiv:2210.14215},
  year={2022}
}

@article{kirkpatrick2017overcoming,
  title={Overcoming catastrophic forgetting in neural networks},
  author={Kirkpatrick, James and Pascanu, Razvan and Rabinowitz, Neil and Veness, Joel and Desjardins, Guillaume and Rusu, Andrei A and Milan, Kieran and Quan, John and Ramalho, Tiago and Grabska-Barwinska, Agnieszka and others},
  journal={Proceedings of the national academy of sciences},
  volume={114},
  number={13},
  pages={3521--3526},
  year={2017},
  publisher={National Academy of Sciences}
}

@article{jang2021towards,
  title={Towards continual knowledge learning of language models},
  author={Jang, Joel and Ye, Seonghyeon and Yang, Sohee and Shin, Joongbo and Han, Janghoon and Kim, Gyeonghun and Choi, Stanley Jungkyu and Seo, Minjoon},
  journal={arXiv preprint arXiv:2110.03215},
  year={2021}
}

@article{olsson2022context,
  title={In-context learning and induction heads},
  author={Olsson, Catherine and Elhage, Nelson and Nanda, Neel and Joseph, Nicholas and DasSarma, Nova and Henighan, Tom and Mann, Ben and Askell, Amanda and Bai, Yuntao and Chen, Anna and others},
  journal={arXiv preprint arXiv:2209.11895},
  year={2022}
}

@article{huang2023mlagentbench,
  title={Mlagentbench: Evaluating language agents on machine learning experimentation},
  author={Huang, Qian and Vora, Jian and Liang, Percy and Leskovec, Jure},
  journal={arXiv preprint arXiv:2310.03302},
  year={2023}
}

@article{jansen2024discoveryworld,
  title={Discoveryworld: A virtual environment for developing and evaluating automated scientific discovery agents},
  author={Jansen, Peter and C{\^o}t{\'e}, Marc-Alexandre and Khot, Tushar and Bransom, Erin and Dalvi Mishra, Bhavana and Majumder, Bodhisattwa Prasad and Tafjord, Oyvind and Clark, Peter},
  journal={Advances in Neural Information Processing Systems},
  volume={37},
  pages={10088--10116},
  year={2024}
}

@article{ha2018world,
  title={World models},
  author={Ha, David and Schmidhuber, J{\"u}rgen},
  journal={arXiv preprint arXiv:1803.10122},
  volume={2},
  number={3},
  pages={440},
  year={2018}
}

@InProceedings{pmlr-v119-dubey20b,
  title = 	 {Kernel Methods for Cooperative Multi-Agent Contextual Bandits},
  author =       {Dubey, Abhimanyu and Pentland, Alex `Sandy'},
  booktitle = 	 {Proceedings of the 37th International Conference on Machine Learning},
  pages = 	 {2740--2750},
  year = 	 {2020},
  editor = 	 {III, Hal Daumé and Singh, Aarti},
  volume = 	 {119},
  series = 	 {Proceedings of Machine Learning Research},
  month = 	 {13--18 Jul},
  publisher =    {PMLR},
  url = 	 {https://proceedings.mlr.press/v119/dubey20b.html}
}

@inproceedings{herrmann2023learning,
  title={Learning one abstract bit at a time through self-invented experiments encoded as neural networks},
  author={Herrmann, Vincent and Kirsch, Louis and Schmidhuber, J{\"u}rgen},
  booktitle={International Workshop on Active Inference},
  pages={254--274},
  year={2023},
  organization={Springer}
}

@techreport{schmidhuber1994learning,
  author      = {Schmidhuber, J{\"u}rgen},
  title       = {On Learning How to Learn Learning Strategies},
  institution = {Fakult{\"a}t f{\"u}r Informatik, Technische Universit{\"a}t M{\"u}nchen},
  number      = {FKI-198-94},
  year        = {1994},
  month       = nov,
  url         = {https://people.idsia.ch/~juergen/FKI-198-94ocr.pdf}
}

@article{schmidhuber1997shifting,
  title={Shifting inductive bias with success-story algorithm, adaptive Levin search, and incremental self-improvement},
  author={Schmidhuber, J{\"u}rgen and Zhao, Jieyu and Wiering, Marco},
  journal={Machine Learning},
  volume={28},
  number={1},
  pages={105--130},
  year={1997},
  publisher={Springer}
}

@book{ring1994continual,
  title={Continual learning in reinforcement environments},
  author={Ring, Mark Bishop},
  year={1994},
  publisher={The University of Texas at Austin}
}

@article{ring1997child,
  title={Child: A first step towards continual learning},
  author={Ring, Mark B},
  journal={Machine Learning},
  volume={28},
  number={1},
  pages={77--104},
  year={1997},
  publisher={Springer}
}

@incollection{schmidhuber2003exploring,
  title={Exploring the predictable},
  author={Schmidhuber, Juergen},
  booktitle={Advances in evolutionary computing: theory and applications},
  pages={579--612},
  year={2003},
  publisher={Springer}
}

@article{elmoznino2026can,
  title={Can In-Context Learning Support Intrinsic Curiosity?},
  author={Elmoznino, Eric and Bhardwaj, Sangnie and von Oswald, Johannes and Nasser, Rajai and Sacramento, Jo{\~a}o and Saurous, Rif A and Lajoie, Guillaume and others},
  journal={arXiv preprint arXiv:2606.19476},
  year={2026}
}

@InProceedings{sancaktar25a,
  title = 	 {{SENSEI}: Semantic Exploration Guided by Foundation Models to Learn Versatile World Models},
  author =       {Sancaktar, Cansu and Gumbsch, Christian and Zadaianchuk, Andrii and Kolev, Pavel and Martius, Georg},
  booktitle = 	 {Proceedings of the 42nd International Conference on Machine Learning},
  pages = 	 {52745--52777},
  year = 	 {2025},
  editor = 	 {Singh, Aarti and Fazel, Maryam and Hsu, Daniel and Lacoste-Julien, Simon and Berkenkamp, Felix and Maharaj, Tegan and Wagstaff, Kiri and Zhu, Jerry},
  volume = 	 {267},
  series = 	 {Proceedings of Machine Learning Research},
  month = 	 {13--19 Jul},
  publisher =    {PMLR},
  url = 	 {https://proceedings.mlr.press/v267/sancaktar25a.html}
}

@article{agarwal2026autodiscovery,
  title={Autodiscovery: Open-ended scientific discovery via bayesian surprise},
  author={Agarwal, Dhruv and Majumder, Bodhisattwa Prasad and Adamson, Reece and Chakravorty, Megha and Gavireddy, Satvika Reddy and Parashar, Aditya and Surana, Harshit and Dalvi Mishra, Bhavana and McCallum, Andrew and Sabharwal, Ashish and others},
  journal={Advances in Neural Information Processing Systems},
  volume={38},
  pages={25181--25219},
  year={2026}
}

% \newpage
% \input{checklist.tex}

\end{document}